\documentclass{article}
\usepackage{PRIMEarxiv}

\usepackage{amsmath,amssymb,amsfonts}
\usepackage{color,soul}
\usepackage{algorithmic}
\usepackage{graphicx}
\usepackage[numbers]{natbib}
\usepackage{subcaption}
\usepackage{textcomp}
\usepackage{xcolor}
\usepackage{hhline}
\usepackage{hyperref}

\def\BibTeX{{\rm B\kern-.05em{\sc i\kern-.025em b}\kern-.08em
    T\kern-.1667em\lower.7ex\hbox{E}\kern-.125emX}}

\title{Improving Variational Autoencoder using Random Fourier Transformation: An Aviation Safety Anomaly Detection Case-Study}

\author{
  Ata Akbari Asanjan \\
  Data Science Group\\
  NASA Ames Research Center\\
  Moffett Field, CA 94305 \\
  Universities Space Research Association \\
  \texttt{ata.akbariasanjan@nasa.gov} \\
  \And
  Milad Memarzadeh \\
  Data Science Group\\
  NASA Ames Research Center\\
  Moffett Field, CA 94305 \\
  \texttt{milad.memarzadeh@nasa.gov} \\
  \And
  Bryan Matthews \\
  Data Science Group\\
  NASA Ames Research Center\\
  Moffett Field, CA 94305 \\
  KBR Inc. \\
  \texttt{bryan.l.matthews@nasa.gov} \\
  \And
  Nikunj Oza \\
  Data Science Group\\
  NASA Ames Research Center\\
  Moffett Field, CA 94305 \\
  \texttt{nikunj.c.oza@nasa.gov} \\
}

\begin{document}
\maketitle

\begin{abstract}
In this study, we focus on the training process and inference improvements of deep neural networks (DNNs), specifically Autoencoders (AEs) and Variational Autoencoders (VAEs), using Random Fourier Transformation (RFT). We further explore the role of RFT in model training behavior using Frequency Principle (F-Principle) analysis and show that models with RFT turn to learn low frequency and high frequency at the same time, whereas conventional DNNs start from low frequency and gradually learn (if successful) high-frequency features. We focus on reconstruction-based anomaly detection using autoencoder and variational autoencoder and investigate the RFT's role. We also introduced a trainable variant of RFT that uses the existing computation graph to train the expansion of RFT instead of it being random. We showcase our findings with two low-dimensional synthetic datasets for data representation, and an aviation safety dataset, called Dashlink, for high-dimensional reconstruction-based anomaly detection. The results indicate the superiority of models with Fourier transformation compared to the conventional counterpart and remain inconclusive regarding the benefits of using trainable Fourier transformation in contrast to the Random variant.
\end{abstract}


\section{Introduction}

Neural networks have emerged as indispensable tools in the realm of artificial intelligence, fueling advancements in image recognition, anomaly detection, and generative modeling, among other domains \cite{de2022review}. Central to their effectiveness is their ability to automatically learn and extract meaningful features from raw data, enabling them to discover complex patterns and relationships. Among the neural architectures, autoencoders stand out for their capacity to diminish dimensionality and capture essential data representations. Autoencoders, first introduced by \cite{rumelhart1985learning}, represent a class of neural network architectures designed for unsupervised learning, dimensionality reduction, and feature learning \cite{baldi2012autoencoders}. Comprising both an encoder and a decoder, autoencoders, specifically the compressive variant commonly referred to as autoencoders, seek to map input data to a lower-dimensional representation and then reconstruct the original data from this compressed form. Autoencoders have found applications in various fields, including anomaly detection, denoising, and data compression, through their ability to learn meaningful feature representations.

Variational Autoencoders (VAEs) have emerged as a significant advancement in generative modeling, extending the capabilities of traditional autoencoders. They effectively model complex data distributions and generate novel, high-quality data samples by learning a probabilistic distribution over the latent space \cite{kingma2019introduction,memarzadeh2020unsupervised,akbari2023probabilistic}. Unlike traditional autoencoders, VAEs incorporate a probabilistic latent space, enabling them to capture uncertainty and generate diverse samples. The reparameterization trick facilitates efficient training by sampling from a simpler distribution and transforming the samples using mean and variance parameters output by the encoder. VAEs typically learn continuous latent representations, enabling smooth interpolation between data points in the latent space. During training, VAEs optimize an objective function that balances a reconstruction loss, measuring fidelity to the input, and a regularization term, typically the Kullback Leibler (KL) divergence between the learned latent distribution and a prior distribution. This encourages the latent space to follow a predefined structure, promoting smoothness and continuity. The probabilistic nature, continuous latent space, and generative modeling capabilities position VAEs as an improvement to autoencoders and make them versatile tools for various data-driven applications, including image synthesis, data augmentation, and robust anomaly detection.

Despite the promising capabilities of neural networks, including autoencoders and VAEs, in learning data with complex patterns, they still struggle with capturing the full complexity and subtlety of the underlying data distributions. Specifically, neural networks struggle with learning high-frequency patterns that often correspond to fine-grained details, subtle variations, and localized features within the data \cite{tancik2020fourier, xu2019frequency, luo2019theory, xu2022overview}. Neural networks, including autoencoders and VAEs, often struggle to represent and model these high-frequency patterns effectively, primarily focusing on the more dominant low-frequency components \cite{xu2019frequency, xu2022overview, luo2019theory}. This behavior of neural networks referred to as "spectral bias", calls for a better understanding of neural network behavior and how it can be mitigated to increase the effectiveness and efficiency of these models. 


Numerous research efforts have been devoted to addressing the high-frequency learning difficulty encountered by neural networks \cite{he2016deep, xu2019frequency, tancik2020fourier}. \citet{he2016deep}  proposed Residual Networks (ResNets) which mitigate the spectral bias by learning the residuals (i.e. high-pass frequencies) instead of the signal in each block. Multiple strategies have been proposed and successfully applied to implicit models, offering a more robust positional encoding technique \cite{zheng2021rethinking,damodaran2023improved} that enhances the model performance. \citet{tancik2020fourier} proposed an integration of sinusoidal mapping techniques applied to the input data. Using the neural tangent kernel (NTK) \cite{jacot2018neural}, the authors illustrated that simple neural networks are impractically slow in learning high-frequency signals in data, and demonstrated significant promise of Random Fourier Transformation in the low-dimensional domain. 

Random Fourier Transformation (RFT) is a versatile technique that leverages principles from Fourier analysis to transform input data into a low-dimensional Euclidean inner product feature space \cite{rahimi2007random}. By introducing the concept of randomness into this transformation, RFT offers a computationally efficient and scalable means of approximating kernel functions. It does so by using random feature maps, thereby enabling the efficient application of kernel methods without explicitly computing kernel matrices. RFT projects the data into a unit circle space where the data's inner product is an unbiased estimator of the kernel. RFT has gained prominence in machine learning, particularly in scenarios where traditional kernel methods, such as the kernel trick in Support Vector Machines, are computationally infeasible due to their high time and memory complexity \cite{rahimi2007random}. RFTs approximate shift-invariant kernels \( k(x, x') = k(x - x') \) which can be viewed as an augmentation technique for shift-prune neural networks architecture (i.e. Dense layers, recurrent layers) to improve their robustness and generalization \cite{vaish2024fourier, xu2023fourier, rahimi2007random}. The impact of RFTs in convolutional blocks is diminished due to convolutional and pooling layers' inherently shift-invariant characteristics. Furthermore, Random Fourier Transformations closely follow the harmonic function concept if all the requirements of orthogonality are respected by the initial randomization \cite{harris1978use, stein1971introduction} which can be regarded as regularization term for neural networks that helps them in the training process.

To overcome the spectral bias limitation and enhance the modeling capabilities of autoencoders and VAEs, we incorporate RFT and leverage the principles of Fourier analysis to expand the input features into a Fourier space. In essence, RFT introduces random samples of frequency domain into the feature space, allowing the models to better capture intricate details and high-frequency patterns in the data. The rationale behind this integration is rooted in the complementary strengths of RFT which is the ability to transform data into a space more amenable to modeling fine-grained details. By incorporating RFT as the precursor layer of the model, we aim to create a novel framework that resolves the spectral bias and learns both low- and high-frequency patterns at the same time.

By further investigating the architecture of RFT, as a precursor layer in neural networks, it becomes obvious that the layer can be thought of as two shared weight random and fixed convolutional layers with a kernel size of 1 that expands the channels of the input data into the pre-defined number of random Fourier channels in parallel with sine and cosine activation functions followed by a concatenation operator. The harmonic activation functions used in this form are also reported to improve the performance of implicit neural networks \cite{sitzmann2020implicit} and neural networks, in general, \cite{liang2021reproducing}. Given the new form of RFT, we described to generate the unbiased kernel, we further explored the possibility of training the kernel for maximum optimization.

By further investigating the architecture of RFT, as a precursor layer in neural networks, it becomes obvious that the layer can be further optimized by allowing relaxing the "random" assumption and allowing the layer to be optimized by the backpropagation. This relaxation transforms the RFT layer into a trainable variant of RFT where the equivalent convolutional layers of kernel size 1 won't suffer from saturation and yield harmonic features of the input data, in which the local and global features are well represented.

This paper empirically investigates the spectral bias issues in neural networks with an emphasis on autoencoders and VAE. It investigates the impacts of random and trainable Fourier transformations on simple data reconstruction as well as reconstruction-based anomaly detection VAEs for complex aviation safety applications.

\section{Methodology}

\subsection{Autoencoder}
\begin{figure}[ht]
    \centering
    \includegraphics[width=0.45\textwidth]{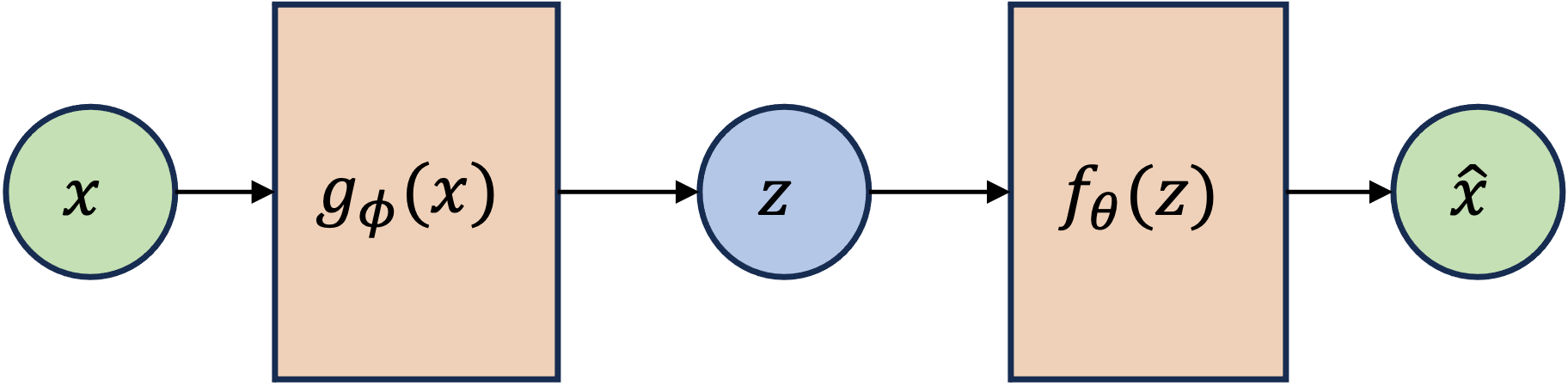}
    \caption{Schematic architecture of autoencoders. Given the input data (\(x\)), latent representation (\(z\)) and reconstructed input data (\(\hat{x}\)), the \(g_{\phi}(.)\), \(f_{\theta}(.)\) represent encoder and decoder components, respectively.}
    \label{fig:fig_ae}
\end{figure}

Autoencoders, introduced by \cite{rumelhart1985learning}, consists of two main components: (1) encoder \((g_{\phi}(.))\): which encodes the data (\(x\)) into latent representation (\(z\)), and (2) decoder \((f_{\theta}(.))\): which transforms latent space back to the input data space \ref{fig:fig_ae}. Autoencoders are mainly categorized into expansive and compressive types, based on larger or smaller latent representations compared to the original data size. Compressive autoencoders are the dominant type due to the constant need for dimensionality reduction, especially while dealing with high-dimensional data. Compressive autoencoders are also considered a form of regularization to prevent "copy-the-input" artifacts when the network learns to copy the input instead of learning its underlying features.

\begin{equation}
    z = g_{\phi}(x)
\end{equation}
\begin{equation}
    \hat{x} = f_{\theta}(z)
\end{equation}

where the encoder and decoder is represented by \((g_{\phi}(.))\) and \((f_{\theta}(.))\), parameterized by \(\phi\) and \(\theta\), respectively. The encoder transforms the data (\(x\)) into latent space representation (\(z\)) and the decoder transforms the latent representation back into a reconstruction of original data (\(\hat{x}\)).

\subsection{Variational Autoencoder}
The latent space in autoencoders is deterministic, representing a fixed transformation of the input data. However, as the need to capture the inherent variability within data became evident, the transition to VAEs emerged. This transition introduces the concept of probabilistic modeling and variational inference into the autoencoder framework \cite{kingma2013auto}. VAEs redefine the latent space as a probability distribution, enabling the encoding process to capture the stochastic nature of data. By incorporating the KL divergence term in their loss function, VAEs strike a balance between data reconstruction and regularization, thereby accommodating encodings with probabilistic variability. 

Variational Inference is the cornerstone for VAEs, allowing them to approximate complex probability distributions in a computationally efficient manner. Variational Inference operates by framing the estimation of a target distribution as an optimization problem, seeking an approximate distribution that best aligns with the true posterior \cite{kingma2013auto, kingma2019introduction}. The proposed training process sidesteps the often intractable direct computation of posterior distributions, a formidable challenge in probabilistic modeling. Consequently, Variational Inference plays a pivotal role in probabilistic graphical models (e.g. VAEs) and enables probabilistic learning of complex data in expressive latent configurations. Mathematically, the goal of Variational Inference is to find the best approximation, often denoted as \(q(\mathbf{z})\), to the true posterior distribution \(p(\mathbf{z} | \mathbf{x})\), where \(\mathbf{z}\) represents latent variables and \(\mathbf{x}\) are the observed data. This approximation is achieved by minimizing the KL divergence between the true posterior and the approximate distribution:

\begin{equation}
    q^{\ast}(\mathbf{z}) = \underset{q(\mathbf{z})}{\text{minimize}} \, \text{KL}\left(q(\mathbf{z}) || p(\mathbf{z} | \mathbf{x})\right)    
\end{equation}

Introducing the KL definition as 

\begin{equation}\label{eq:KL_eq}
    \begin{split}
        \text{KL}(q(z) || p(z | x)) &= \int q(z) \log\left(\frac{q(z)}{p(z | x)}\right) \, dz \\
        &= \mathbb{E}_{q(z)}\left[\log q(z)\right] - \mathbb{E}_{q(z)}\left[\log p(z | x)\right] \\
        &= \mathbb{E}_{q(z)}\left[\log q(z)\right] - \mathbb{E}_{q(z)}\left[\log p(z, x)\right] + \log p(x)
    \end{split}
\end{equation}

where the \( \log p(x) \) is often intractable and cannot be calculated. This optimization problem is often solved by defining a lower bound instead of an exact solution and maximizing the Evidence Lower Bound (ELBO), also known as the variational objective (Eq. (\ref{eq:ELBO_2})), which is given by:

\begin{equation}\label{eq:ELBO_1}
    -\text{KL}(q(z) || p(z | x)) \geq \mathbb{E}_{q(z)}\left[\log p(z, x)\right] - \mathbb{E}_{q(z)}\left[\log q(z)\right]
\end{equation}
\begin{equation}\label{eq:ELBO_2}
    \text{ELBO} = \mathbb{E}_{q(z)}\left[\log p(z, x)\right] - \mathbb{E}_{q(z)}\left[\log q(z)\right]
\end{equation}
In the context of VAEs, Variational Inference plays a pivotal role in approximating the posterior distribution of the latent variables \( (q_{\phi}(z|x)) \). This occurs during the utilization of the Kullback-Leibler (KL) divergence, as defined in Eq. (\ref{eq:KL_eq}), to determine the most accurate approximation to the genuine posterior distribution:

\begin{equation}\label{eq:KL_eq_2}
    \begin{split}
        - \text{KL}\left( q_\phi(z | x) \, || \, p(z | x) \right) \geq\\
        \mathbb{E}_{q_\phi(z | x)} \left[ \log\left(p_\theta(x | z) p_\theta(z)\right) - \log q_\phi(z | x) \right] 
    \end{split}
\end{equation}

Eq. \ref{eq:KL_eq_2} is the leading equation to the VAE loss term and can be further simplified as:

\begin{equation}
    \mathcal{L}(\phi, \theta; x) = \mathbb{E}_{q_{\phi}(z|x)}\left[\log p_{\theta}(x|z)\right] - \text{KL}\left(q_{\phi}(z|x) || p_{\theta}(z)\right)
\end{equation}

\begin{figure}[ht]
    \centering
    \includegraphics[width=0.45\textwidth]{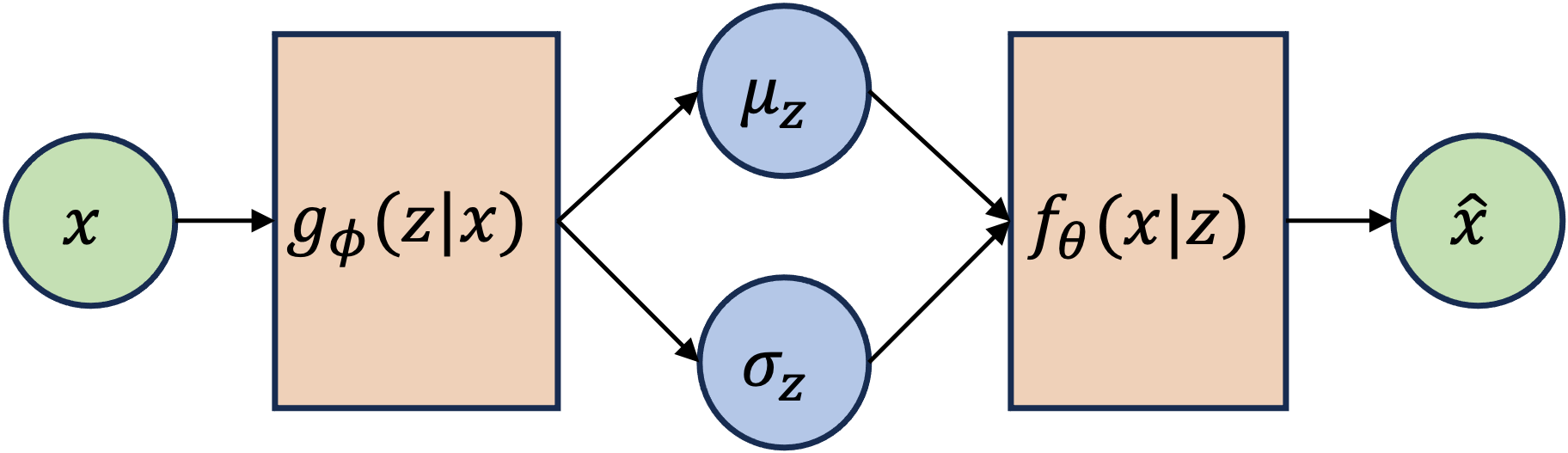}
    \caption{Schematic architecture of autoencoders. Given the input data (\(x\)), latent representation (\(z\)) and reconstructed input data (\(\hat{x}\)), the \(g_{\phi}(.)\), \(f_{\theta}(.)\) represent encoder and decoder components, respectively.}
    \label{fig:fig_vae}
\end{figure}

\subsection{Random \& Trainable Fourier Transformation}
The essence of RFT is grounded in Fourier transformation. Given a data point \(x\), RFT maps it to a new feature space using random linear combinations of sinusoidal basis functions. This transformation is realized by projecting the data into the space of the Fourier coefficients. The general trigonometric mathematical representation of the RFT can be succinctly expressed as follows:

\begin{equation}
   \gamma(x) = \begin{bmatrix} a_1 \cos(2\pi b_1^T x) \\ a_1 \sin(2\pi b_1^T x) \\ \vdots \\ a_m \cos(2\pi b_m^T x) \\ a_m \sin(2\pi b_m^T x) \end{bmatrix} 
\end{equation}

The given function represents the mapping of input \(x\) to the transformed space \(\gamma(x)\) using trigonometric terms, where \(a_i\) and \(b_i\) (for \(i = 1, 2, ..., m\)) are model parameters. Here, \(m\) denotes the number of Fourier features used in the transformation. These parameters can either be fixed values or trained using backpropagation via a computation graph. In many cases, the parameter \(a_i\) is set to one because the neural network following Random Fourier Transformation (RFT) can automatically assign weights to each component, effectively reducing \(\gamma(x)\) to combinations of \(sin(\cdot)\) and \(cos(\cdot)\) functions with only \(b_i\) (for \(i = 1, 2, ..., m\)) as the RFT parameters. This simplification aids in computational efficiency and the interpretability of the transformed features within the neural network architecture.

This equation represents the mapping of input \(x\) to the transformed space '\(\gamma(x)\)' using trigonometric terms, where \(a_{i}\) and \(b_{i}\) (for \(i = 1, 2, ..., m\)) are model parameters that can be either fixed or trained using backpropagation via computation graph. Commonly, the parameter \(a\) is set to constant as the neural network following RFT can assign weights to each component automatically and reduce the \(\gamma(x)\) to the combinations of \(sin(.)\) and \(cos(.)\) with only \(b\) as the RFT parameters.

Trainable Fourier Transformation is a variation of RFT, in which the parameters \( b_i \) are optimized through gradient-based training within the computation graph. This relaxation in the RFT assumption transforms the layer into a trainable channel-wise neural network layer.
\section{Datasets}

\subsection{Low Dimensional Datasets}

In investigating spectral bias in neural network learning, we generated two distinct datasets to explore the impact of dimensional complexity on model performance. The first dataset is a low-dimensional set comprising 10,000 samples, each with values ranging between -1, 0, and 1. To introduce a threshold for this dataset, we set the threshold value at 0.5, which serves as a boundary for classification tasks. This dataset allows us to examine how neural networks learn and generalize from simple, discrete input patterns, shedding light on any spectral biases that may arise due to the limited dimensional complexity.

In addition to the low-dimensional dataset, we constructed a one-dimensional dataset using the equation \( y = \sin(x) + \sin(3x) + \sin(5x) \). This equation generates a sinusoidal wave with multiple frequencies, resulting in a more intricate and continuous input pattern compared to the discrete values in the low-dimensional dataset. By leveraging this one-dimensional dataset, we aim to evaluate how neural networks handle varying spectral properties and whether biases emerge in learning representations of more complex, continuous data. Together, these datasets provide a comprehensive framework for analyzing spectral bias in neural network learning across different levels of input dimensionality and complexity, facilitating a deeper understanding of model behavior and performance in real-world applications.

The benchmarks used in this study are adapted from \cite{xu2019frequency} to illustrate how neural networks perform when trained on straightforward datasets. These benchmarks serve as a means to evaluate the behavior and capabilities of neural networks when faced with simplified data structures. By leveraging these benchmarks, we aim to gain insights into how neural networks handle and learn from basic datasets, providing a foundational understanding of their performance in more complex scenarios. The utilization of established benchmarks also allows for comparisons with existing research, facilitating a deeper analysis of neural network behavior across different experimental setups and dataset complexities.

\subsection{High Dimensional Datasets}\label{subsect:high_dim}

The dataset used for this study focuses on detecting anomalies related to the deployment of flaps and path deviations during the final approach to landing in commercial aviation \cite{memarzadeh2022multiclass}. The anomalies in the dataset are characterized by a delay in flap deployment, lasting approximately 160 seconds, as identified by subject matter experts. It's important to note that while this labeled anomaly is included in the dataset, it is not utilized as an input during unsupervised training. The dataset consists of ten time-series representing aeronautical sensor outputs, encompassing parameters such as aircraft position, orientation, speed, and the positions of control surfaces.

This dataset captures a 160-second window snapshot of 16,000 flights, a subset of the larger 99,000 flights dataset, during the final approach phase, specifically when the flights are crossing the 1,000-foot mark before touchdown. The dataset includes three different types of anomalies related to flap deployment, path deviations, and speed anomalies, alongside a nominal class representing normal flight behavior. The data originates from modern commercial aircraft equipped with flight data recorders that record various discrete and continuous parameters at approximately 1 Hz throughout the flight duration. These parameters encompass flight control systems, actuators, engines, landing gear, avionics, pilot commands, and other relevant aspects of aircraft operations.

The data mining and knowledge discovery process applied to this dataset involves scalable multiple-kernel learning techniques for large-scale distributed anomaly detection. A novel multivariate time-series search algorithm is employed to search for signatures of discovered anomalies within massive datasets. The goal is to identify operationally significant events related to environmental, mechanical, and human factors that may impact aviation safety. The process is designed to complement traditional human-generated analysis methods and aims to discover previously unknown aviation safety incidents. The dataset is meticulously validated by independent domain experts to ensure the accuracy and reliability of the discovered anomalies.

\section{Results and Discussion}

Our results are split into synthesized low-dimensional benchmark datasets and real-world aviation safety datasets. In our synthesized benchmark datasets we seek to shed light on the learning mechanism of RFT and TFT in simple feedforward neural networks. Then, we extend our understanding to autoencoders and variational autoencoders designed to learn the nominal aviation dataset and recognize the anomalous events in inference based on higher reconstruction errors in the test set. It is important to note that to maintain consistency in the comparison, we duplicate the input data for both RFT and TFT models. This ensures that all cases have an equal number of inputs and model parameters.

\subsection{Benchmark Experiments}
\begin{figure}[h]
    \centering
    \begin{subfigure}[b]{\columnwidth}
        \includegraphics[width=\textwidth]{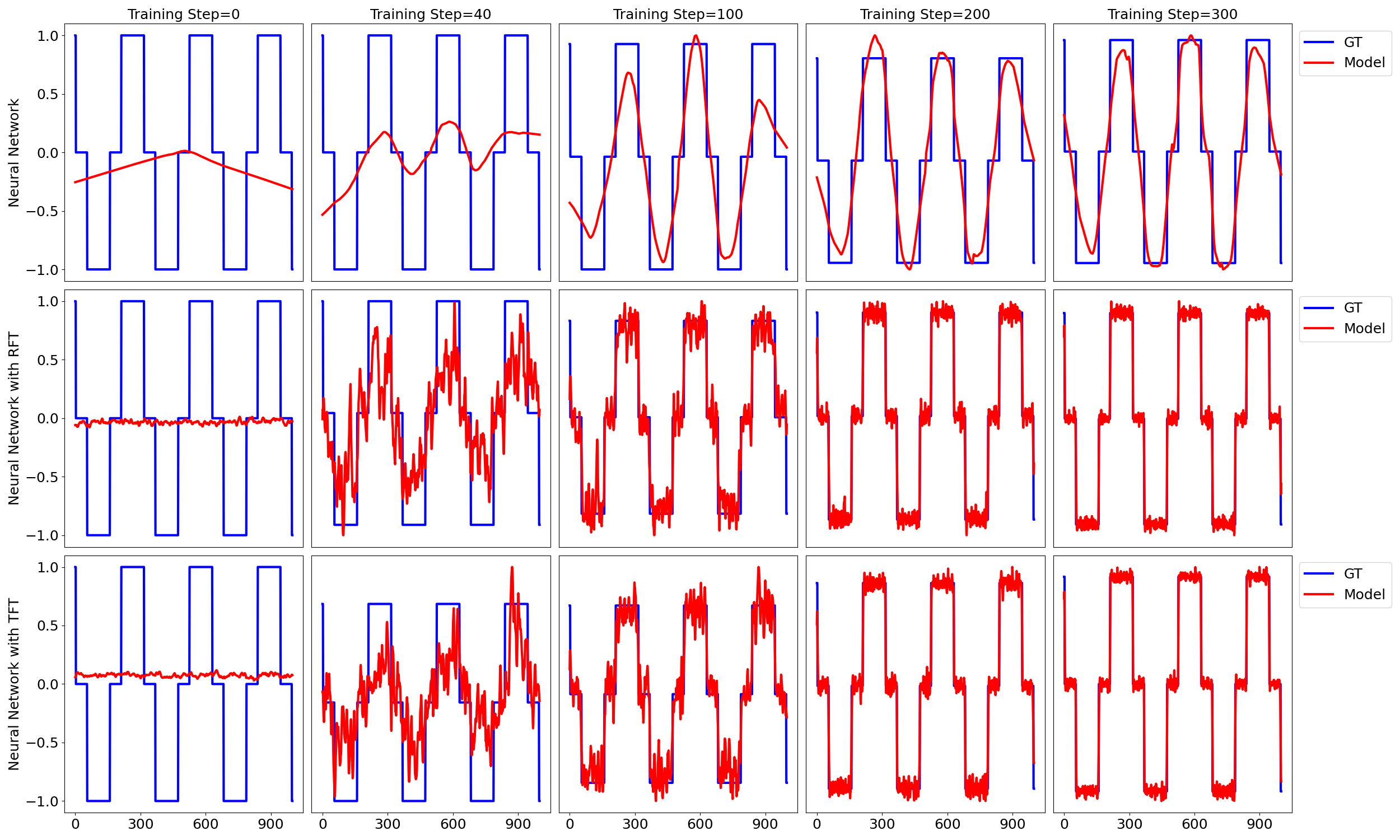}
        \caption{Step Function}
        \label{fig:signal_step}
    \end{subfigure}
    \quad
    \begin{subfigure}[b]{\columnwidth}
        \includegraphics[width=\textwidth]{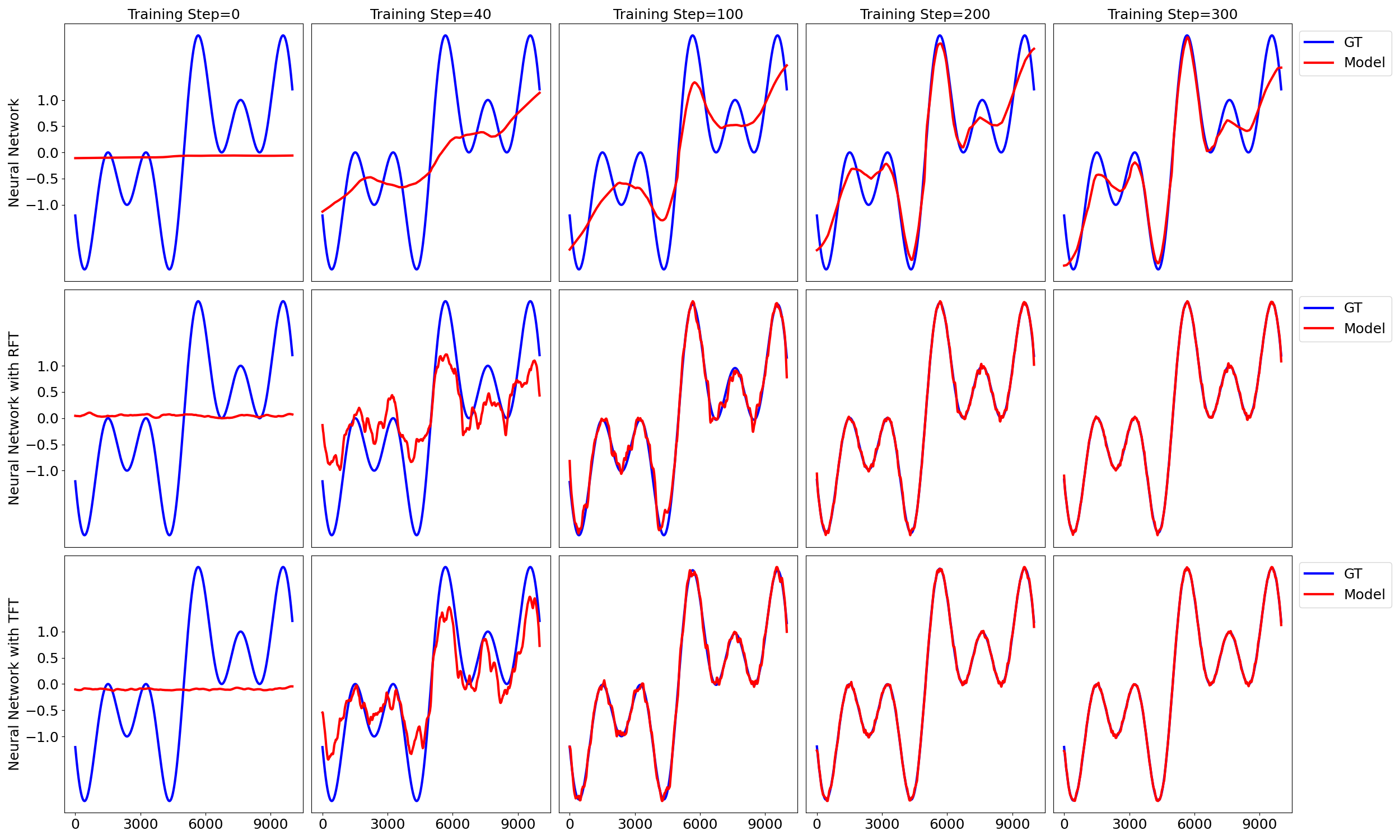}
        \caption{Mixture of Sines Function}
        \label{fig:signal_sine}
    \end{subfigure}
    \caption{Comparison of model fitting performance using vanilla neural networks, neural networks with Trainable Fourier transformations, and neural networks with Trainable Fourier Transformations on both Step and mixture of Sines datasets (from top row to bottom). Plots depict the convergence of models (red line) to ground truth (blue line) across increasing training epochs, from left to right}
    \label{fig:signals}
\end{figure}

For the benchmark experiment, we demonstrate the model evolution through training epochs and how the model is learning the signals. Fig. \ref{fig:signals} illustrates the performance of three neural network architectures on both datasets, namely the vanilla neural network, neural network with Random Fourier Transformation, and neural network with Trainable Fourier Transformation. Across the top row to the bottom row, the plots showcase the model fitting process, with the red line representing the model's fit to the ground truth data shown by the blue line. The results indicate that both the neural network with Random Fourier Transformation and the neural network with Trainable Fourier Transformation exhibit faster convergence to the data compared to the vanilla neural network, particularly in regions with sharp features. Interestingly, there are no significant differences observed between the Random Fourier and Trainable Fourier components, suggesting minimal gains by training the parameters in the Fourier Transformation process.

In our exploration of model training dynamics, we investigate the Frequency-Principle analysis as proposed by \citeauthor{xu2019frequency} \cite{xu2019frequency}. This analysis method, detailed in their work, involves decomposing signals into Fourier frequencies. By employing this approach, we aim to gain deeper insights into the spectral bias phenomenon prevalent in neural network learning processes. Spectral bias refers to the tendency of neural networks to favor learning certain frequency components of input data over others, which can significantly influence model performance and generalization.

\citeauthor{xu2019frequency}'s \cite{xu2019frequency} Frequency-Principle analysis serves as a valuable framework for dissecting how neural networks interact with and prioritize different frequency components during training. Through this analysis, we seek to uncover patterns and tendencies in model behavior related to frequency-specific learning preferences. By understanding these dynamics, we can make informed adjustments to training strategies, network architectures, or preprocessing techniques to mitigate spectral bias effects and enhance overall model robustness and performance across various datasets and tasks.

\begin{figure}[h]
    \centering
    \begin{subfigure}[b]{\columnwidth}
        \includegraphics[width=\textwidth]{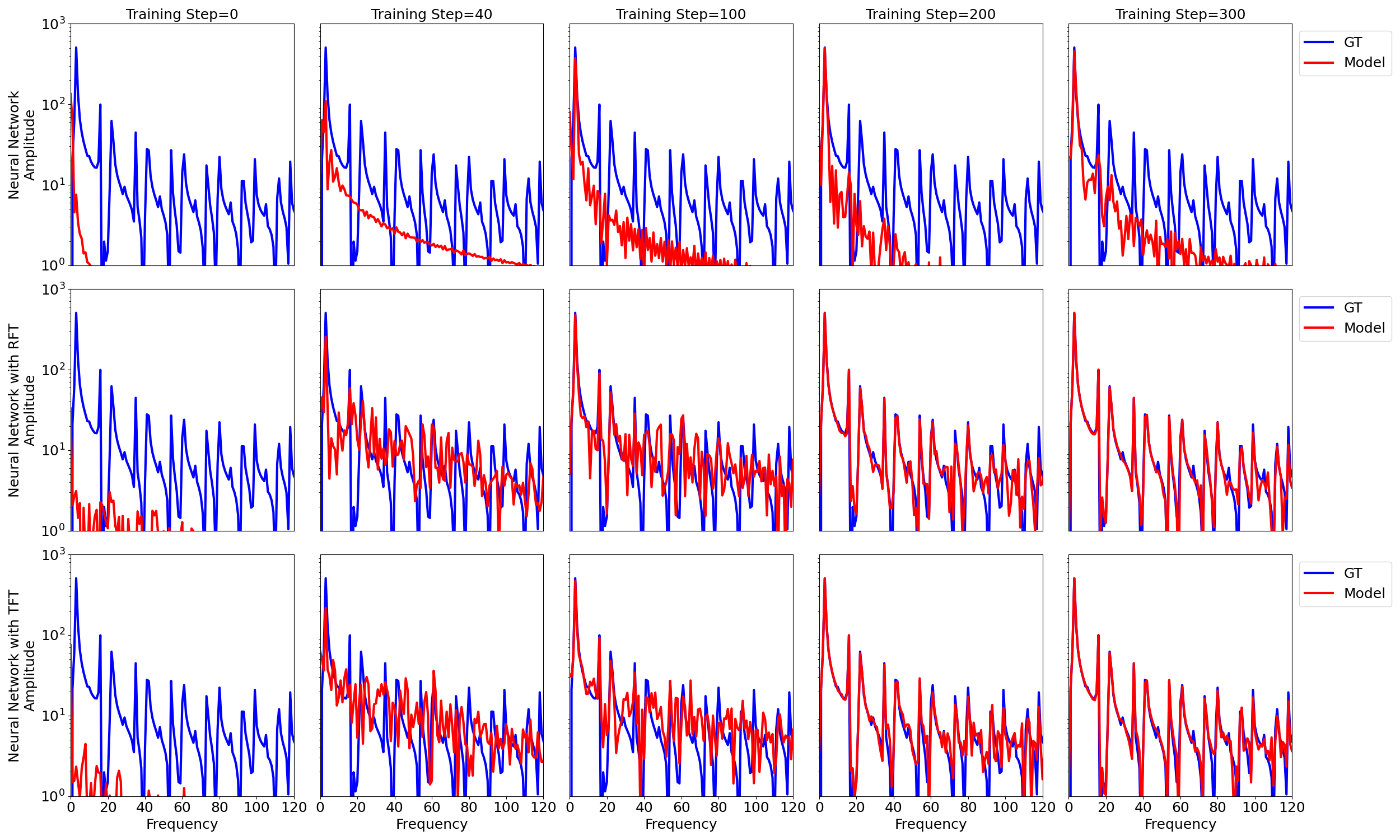}
        \caption{Step Function}
        \label{fig:freq_step}
    \end{subfigure}
    \quad
    \begin{subfigure}[b]{\columnwidth}
        \includegraphics[width=\textwidth]{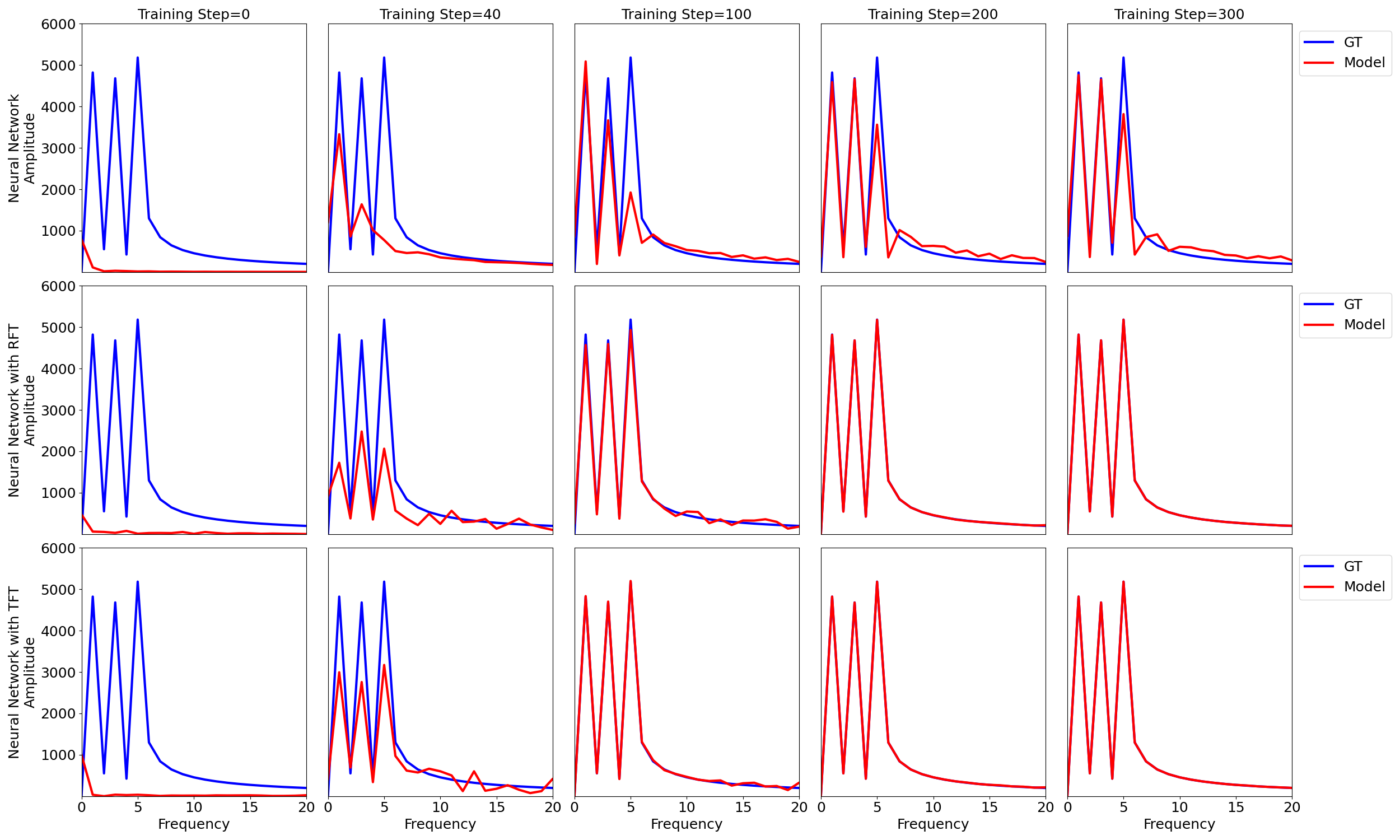}
        \caption{Mixture of Sines Function}
        \label{fig:freq_sine}
    \end{subfigure}
    \caption{With a similar visualization structure to Fig. \ref{fig:signals}, frequency analysis results showcase the learning behavior of neural networks with and without Fourier transformations. The figure illustrates that while the vanilla neural network learns low-pass frequencies before high-pass frequencies, the neural networks with Random Fourier Transformation and Trainable Fourier Transformation exhibit simultaneous learning of both low and high frequencies.}
    \label{fig:freqs}
\end{figure}

In our study, we conducted a frequency analysis to examine how neural networks learn different frequency components within a signal. Fig. \ref{fig:freqs} presents the results of this analysis, highlighting the distinct learning patterns observed in neural networks with various architectures. The vanilla neural network demonstrates a tendency to learn low-pass frequencies earlier in the training process, gradually incorporating high-pass frequencies as training progresses. This behavior aligns with the known spectral bias phenomenon, where neural networks prioritize certain frequency components over others. However, contrasting results are observed in the neural networks with Random Fourier Transformation and Trainable Fourier Transformation. These networks show the ability to learn both low and high frequencies simultaneously, indicating mitigation of spectral bias effects. The simultaneous learning of frequency components suggests that Fourier transformations, whether random or trainable, can help neural networks achieve a more balanced representation of signal characteristics, potentially enhancing model performance and generalization capabilities.

\subsection{Reconstruction-based Anomaly Detection}

\begin{figure}[ht]
    \centering
    \includegraphics[width=\columnwidth]{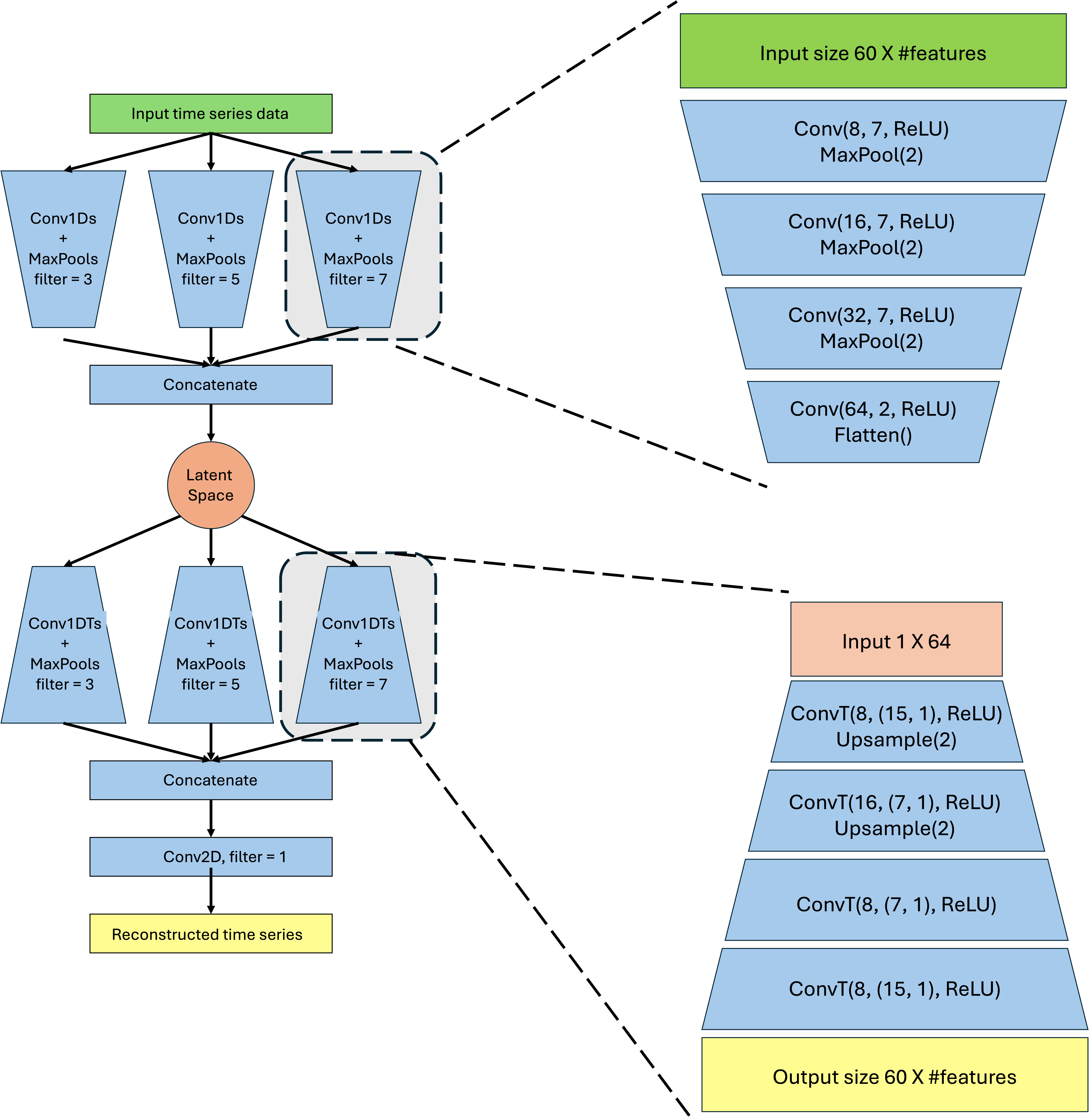}
    \caption{Detailed model architecture for CVAE and CAE models. The primary distinction between CVAE and CAE lies in their respective methods of calculating the "Latent Space." CVAE achieves this by defining multivariate Gaussian distribution parameters (\( \mu \) and \( \sigma \)), followed by sampling from the distribution. Conversely, CAE directly computes the latent representation.}
    \label{fig:model_arch}
\end{figure}

We further analyze the behavior of RFT and TFT in a more complex dataset and neural network architecture by employing the high-dimensional time-series dataset introduced in \ref{subsect:high_dim}. The data originally has 10 variables which are transformed into 64 Fourier features before entering the main autoencoder and Convolutional VAE (CVAE models. For this experiment, we use a similar architecture for both VAE and autoencoder (from now on referred to as CAE) which can be found in \ref{fig:model_arch}. The only difference is the latent space of CVAE will represent an encoded representation instead of the latent distribution and thus, is not trained to minimize ELBO and only focuses on reducing reconstruction error.

\begin{figure}
    \centering
    \includegraphics[width=\columnwidth]{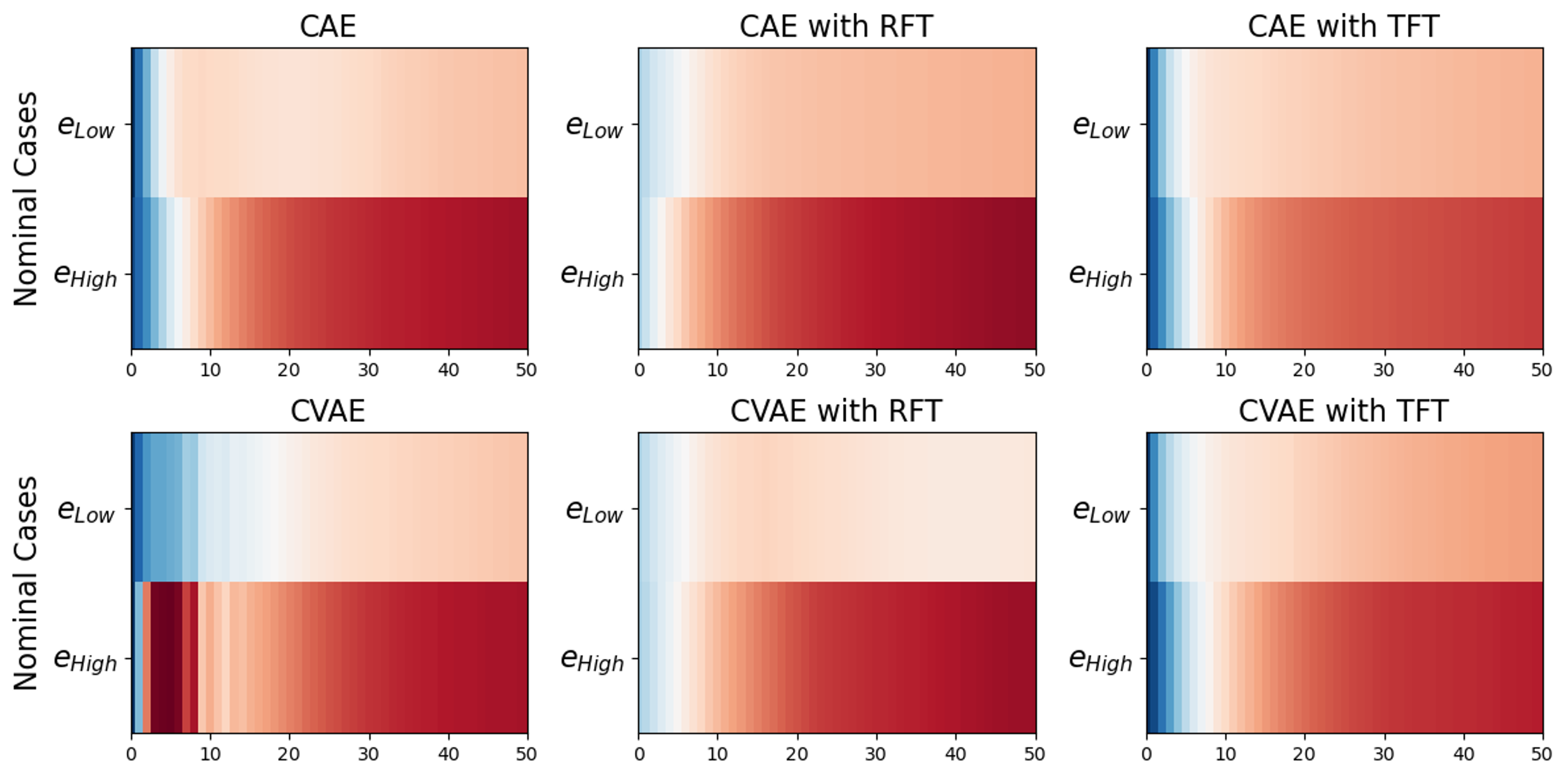}
    \caption{The frequency behavior of CAE, CAE with RFT, and CAE with TFT (first row) for the nominal of the Flaps anomaly dataset. Similar plots follow from the third row for CVAE, CVAE with RFT, and CVAE with TFT (second row). The y-axis in each plot represents the separation of low and high frequencies and the x-axis shows the timesteps of training evolution.}
    \label{fig:flaps}
\end{figure}

Figure \ref{fig:flaps} illustrates the learning behavior of CAE (first row) and CVAE (second row) across different configurations: without additional components, with RFT, and with TFT integrated into the original model for Flaps anomaly. In CAE, the results demonstrate enhanced high-pass filter learning with the inclusion of RFT and TFT, albeit with minimal improvements. On the other hand, CVAE exhibits a noisy learning paradigm on its own but achieves smooth and balanced learning when considering both low and high frequencies together. This balanced learning pattern is similarly observed in CVAE when combined with TFT.

\begin{figure}
    \centering
    \includegraphics[width=\columnwidth]{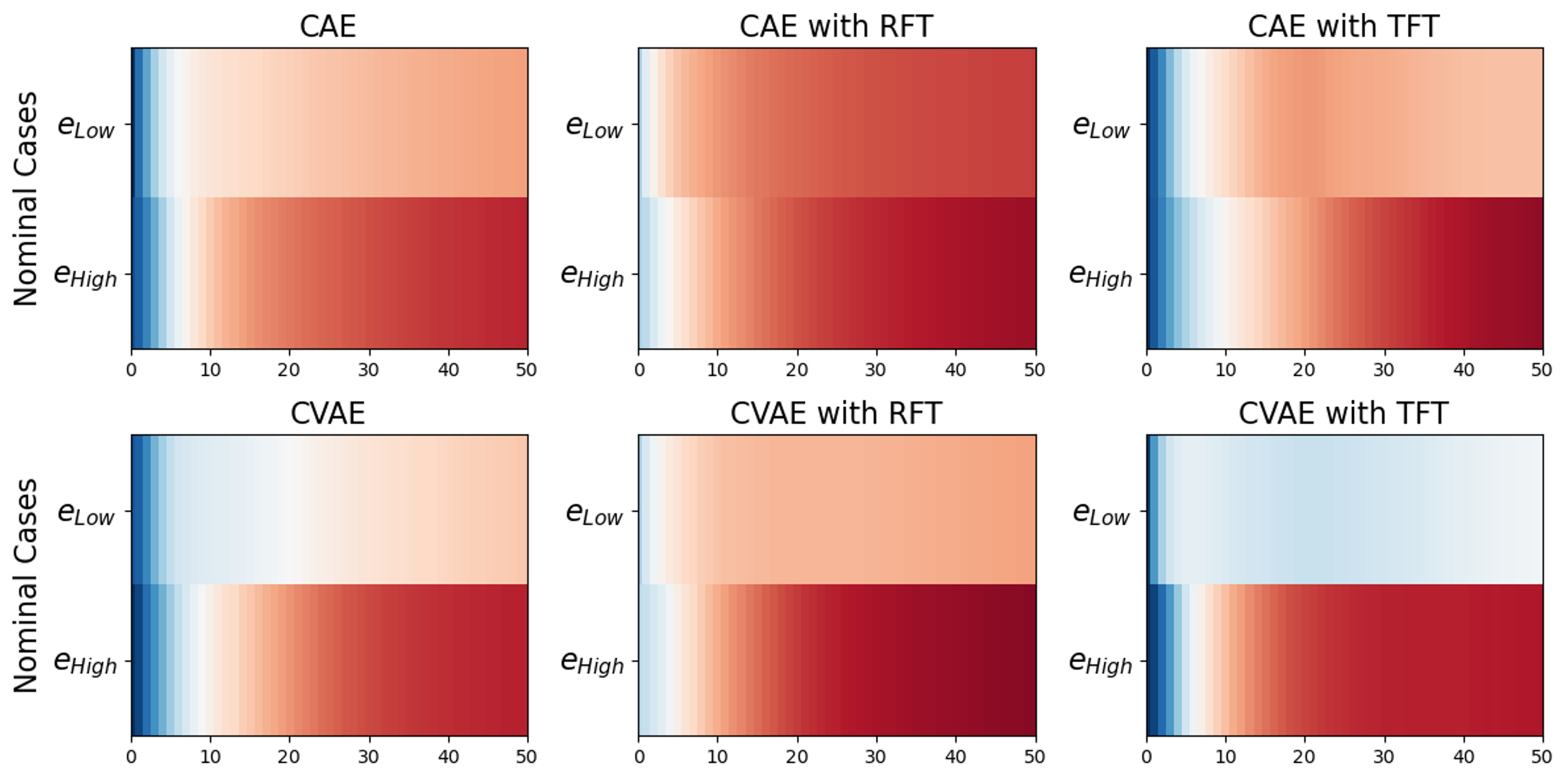}
    \caption{The frequency behavior of CAE, CAE with RFT, and CAE with TFT (first row) for the nominal of the Path anomaly dataset. Similar plots follow from the third row for CVAE, CVAE with RFT, and CVAE with TFT (second row). The y-axis in each plot represents the separation of low and high frequencies and the x-axis shows the timesteps of training evolution.}
    \label{fig:path}
\end{figure}

Figure \ref{fig:path} presents the learning behavior of the Path anomaly dataset, akin to the Flaps dataset. In line with the Flaps results, CAE initially learns low-pass filters before proceeding to high-pass filters. However, when integrated with RFT, CAE shows accelerated learning across both frequencies, achieving a balanced outcome. Conversely, CAE with TFT demonstrates limitations in effectively learning the frequencies compared to CAE with RFT.
In the case of CVAE models, integrating RFT enhances learning behavior akin to CAE's, while TFT does not yield similar improvements.

\begin{figure}
    \centering
    \includegraphics[width=\columnwidth]{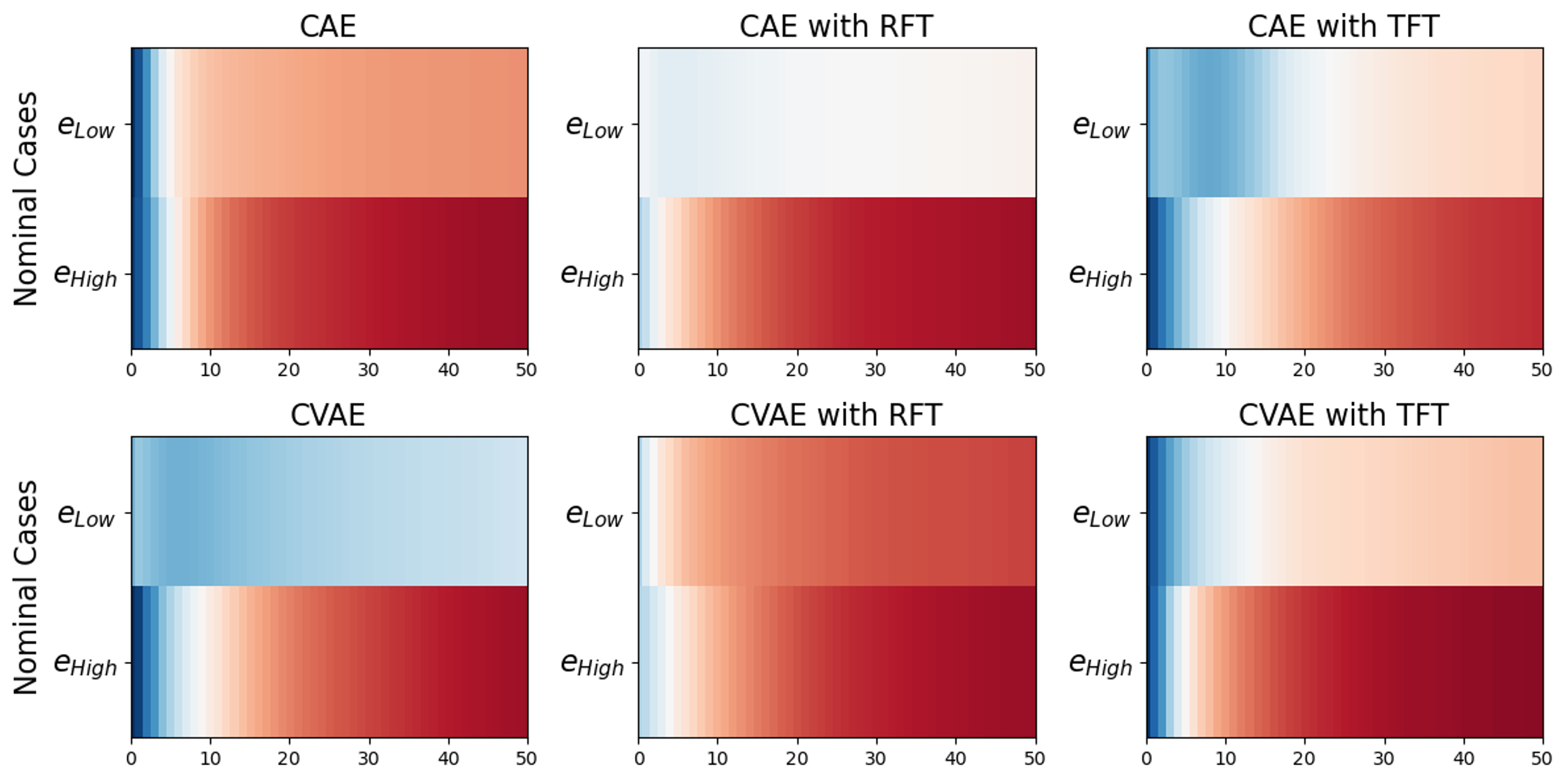}
    \caption{The frequency behavior of CAE, CAE with RFT, and CAE with TFT (first row) for the nominal of the Speed anomaly dataset. Similar plots follow from the third row for CVAE, CVAE with RFT, and CVAE with TFT (second row). The y-axis in each plot represents the separation of low and high frequencies and the x-axis shows the timesteps of training evolution.}
    \label{fig:speed}
\end{figure}

In Figure \ref{fig:speed}, we observe CAE's performance in learning the Speed dataset, which is comparatively fair. However, CAE with RFT demonstrates better learning of high-pass filters but struggles with converging on low-pass features. CAE with TFT shows slightly improved performance compared to CAE with RFT but still falls short of CAE's overall performance.

On the contrary, CVAE faces challenges in learning the low-pass features independently. However, integrating RFT into CVAE results in effective learning of both low and high-pass features. While CVAE with TFT enhances CVAE's performance, it does not reach the level of improvement seen with CVAE and RFT.

In terms of metric performances, we compare the results in terms of precision, recall, and F1-score. In order to eliminate the training and inference variability, we train and infer from each model for 10 independent runs, and the results in Tables \ref{tab:Flaps}-\ref{tab:Speed} are the average of the statistics for the 10 runs.

For the Flaps anomaly (Table \ref{tab:Flaps}), the results highlight the superior performance of RFT and TFT models within both CAE and CVAE frameworks. A comparison between CAE and CVAE results reveals the CVAE's improved performance, attributed to its enhanced model architecture. However, with the introduction of Fourier transformation (both RFT and TFT), the performance gap between CAE and CVAE diminishes, especially in terms of Recall and F1-score. Notably, there is no significant performance gain observed between RFT and TFT, indicating that either RFT adequately represents the data or TFT lacks the capacity to further enhance the initialized Fourier transformations.

\begin{table}[h]
    \centering
    \begin{tabular}{|c|c|c|c|}
        \hline
         {} & Precision & Recall & F1-score \\
         \hhline{|=|=|=|=|}
         {CAE} & 50.25 & 53.51 & 51.83 \\
         \hline
         {CAE with RFT} & 59.02 & \textbf{84.86} & 69.62 \\
         \hline
         {CAE with TFT} & 62.5 & 83.78 & \textbf{71.59} \\
         \hline
         {CVAE} & 62.15 & 59.46 & 60.77 \\
         \hline
         {CVAE with RFT} & \textbf{63.32} & 78.38 & 70.05 \\
         \hline
         {CVAE with TFT} & 60.91 & 72.43 & 66.17 \\
         \hline
    \end{tabular}
    \caption{Performance metrics (Precision, Recall, and F1-score) of CAE, CAE with RFT, CAE with TFT, CVAE, CVAE with RFT, and CVAE with TFT models evaluated on the Flaps anomaly dataset.}
    \label{tab:Flaps}
\end{table}

For Path anomaly (Table \ref{tab:Path}), we observe a similar performance improvement by incorporating CVAE instead of CAE. Similar to the Flaps anomaly, using RFT and TFT tightens the performance gap between the two models, The results further indicate that in the case of Path anomaly, CVAE with RFT significantly outperforms the TFT variant. The result from CAE demonstrates a similar pattern, showing the CAE with RFT outperforms the TFT variant. This further indicates that gradient-based training does not guarantee an improved Fourier transformation.

In the Path anomaly analysis (Table \ref{tab:Path}), we note a comparable performance enhancement when utilizing CVAE over CAE. As observed in the Flaps anomaly, the inclusion of RFT and TFT narrows the performance difference between these models in Recall and F1-score metrics. Particularly noteworthy is the superior performance of CVAE with RFT compared to its TFT counterpart in addressing the Path anomaly. This trend is mirrored in the CAE results, where CAE with RFT surpasses the TFT variant. These findings underscore that merely relying on gradient-based training methods does not inherently ensure an optimized Fourier transformation in the case of Path anomaly dataset.

\begin{table}[]
    \centering
    \begin{tabular}{|c|c|c|c|}
        \hline
         {} & Precision & Recall & F1-score \\
         \hhline{|=|=|=|=|}
         {CAE} & 39.13 & 10.54 & 16.61 \\
         \hline
         {CAE with RFT} & 57.21 & 30.68 & 39.94 \\
         \hline
         {CAE with TFT} & 55.96 & 28.57 & 37.83 \\
         \hline
         {CVAE} & 40.98 & 11.71 & 18.21 \\
         \hline
         {CVAE with RFT} & \textbf{58.87} & \textbf{30.91} & \textbf{40.49} \\
         \hline
         {CVAE with TFT} & 55.47 & 26.5 & 35.88 \\
         \hline
    \end{tabular}
    \caption{Performance metrics (Precision, Recall, and F1-score) of CAE, CAE with RFT, CAE with TFT, CVAE, CVAE with RFT, and CVAE with TFT models evaluated on the Path anomaly dataset.}
    \label{tab:Path}
\end{table}

\begin{table}[]
    \centering
    \begin{tabular}{|c|c|c|c|}
        \hline
         {} & Precision & Recall & F1-score \\
         \hhline{|=|=|=|=|}
         {CAE} & 50.05 & 4.99 & 9.07 \\
         \hline
         {CAE with RFT} & 54.19 & 7.52 & 13.21 \\
         \hline
         {CAE with TFT} & \textbf{54.31} & \textbf{8.61} & \textbf{14.87} \\
         \hline
         {CVAE} & 51.03 & 5.06 & 9.2 \\
         \hline
         {CVAE with RFT} & 53.16 & 6.9 & 12.22 \\
         \hline
         {CVAE with TFT} & 54.22 & 8.34 & 14.45 \\
         \hline
    \end{tabular}
    \caption{Performance metrics (Precision, Recall, and F1-score) of CAE, CAE with RFT, CAE with TFT, CVAE, CVAE with RFT, and CVAE with TFT models evaluated on the Speed anomaly dataset.}
    \label{tab:Speed}
\end{table}

The table presents the performance metrics (Precision, Recall, and F1-score) of different models evaluated on the Speed anomaly dataset. 

For CAE, we observe Precision at 50.05, Recall at 4.99, and F1-score at 9.07. Moving to CAE with RFT, we see improvements in Precision (54.19), Recall (7.52), and F1-score (13.21). Similarly, CAE with TFT exhibits further enhancements in Precision (54.31), Recall (8.61), and F1-score (14.87), showcasing the impact of incorporating temporal and frequency transformations.

Comparing the CAE variants to their CVAE counterparts, we notice a slight increase in performance. CVAE shows Precision at 51.03, Recall at 5.06, and F1-score at 9.2. However, the addition of RFT boosts these metrics for CVAE with RFT (Precision: 53.16, Recall: 6.9, F1-score: 12.22) and CVAE with TFT (Precision: 54.22, Recall: 8.34, F1-score: 14.45).

The results suggest that incorporating RFT and TFT can significantly improve model performance across various metrics, particularly evident in the comparison between CAE and CVAE models with these transformations. However, it is difficult to recognize a significant benefit of the trainable Fourier transformation (TFT).

Additionally, we analyzed the training behavior of individual variables within both the nominal and anomaly data partitions in the test set. Comparing the performance of each model helps us gain insights into what distinguishes an anomaly as an outlier sample for each model. We focused on comparing CAE and CAE with RFT for simplicity, so we are comparing the best performance with its closest model.

\begin{figure}
    \centering
    \includegraphics[width=\columnwidth]{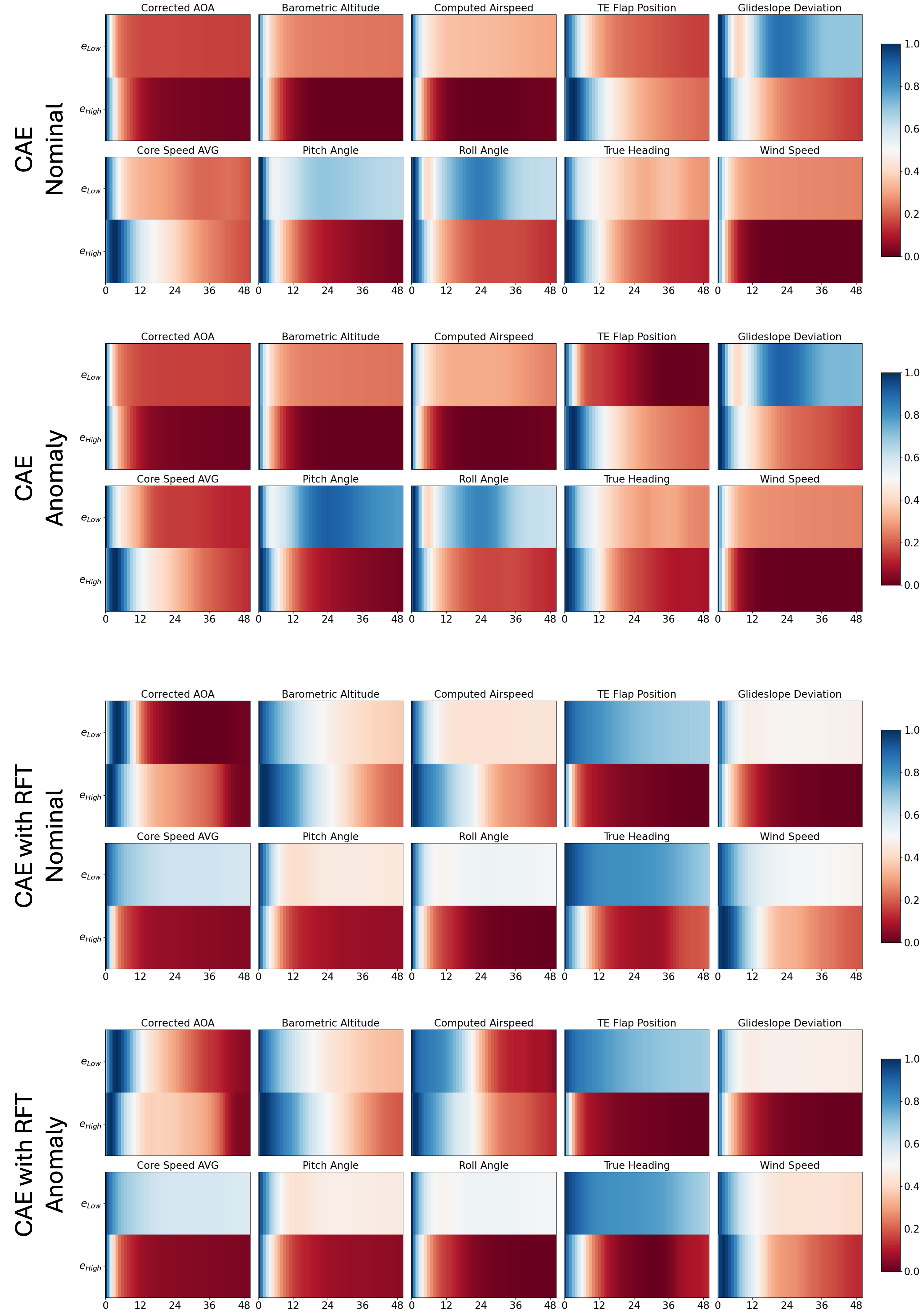}
    \caption{The frequency behavior of CAE and CAE with RFT on nominal and anomaly segments for each variable for Flaps anomaly. Each subplot illustrates the separation of low and high frequencies on the y-axis, while the x-axis indicates the training evolution over time.}
    \label{fig:flaps_var}
\end{figure}

In CAE, the nominal and anomaly sets exhibit similarity across all variables, with only a minor visible difference observed in Computed airspeed (see Figure \ref{fig:flaps_var}). On the other hand, CAE with RFT shows more noticeable discrepancies, particularly in Computed Airspeed and Computed Angle of Attack.

\begin{figure}
    \centering
    \includegraphics[width=\columnwidth]{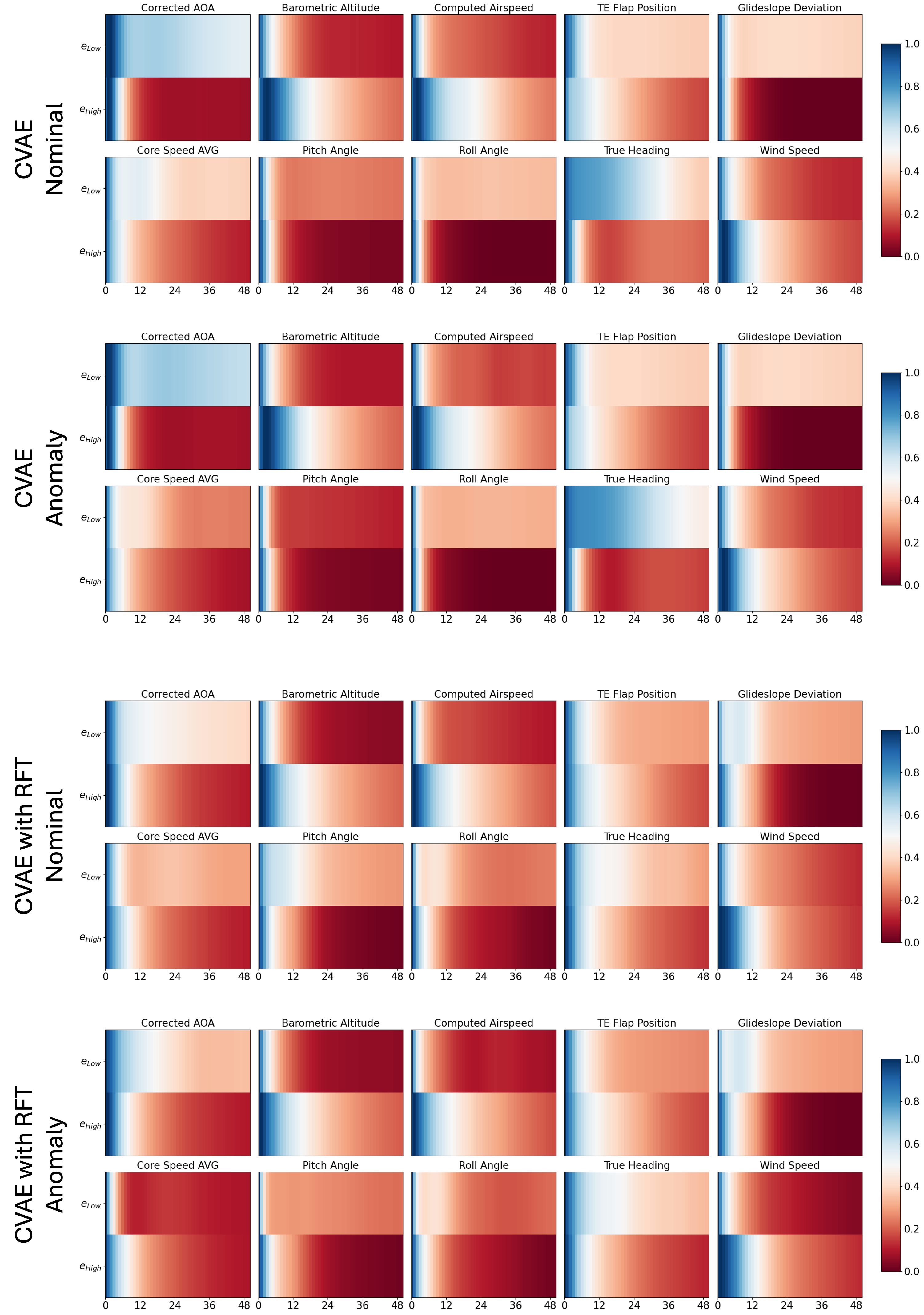}
    \caption{The frequency behavior of CAE and CAE with RFT on nominal and anomaly segments for each variable for Path anomaly. Each subplot illustrates the separation of low and high frequencies on the y-axis, while the x-axis indicates the training evolution over time.}
    \label{fig:path_var}
\end{figure}

Figure \ref{fig:path_var} follows a similar approach to Figure \ref{fig:flaps_var}, where we juxtapose the top-performing model against its base variant for comprehensive analysis. By examining the nominal and anomaly plots across all variables, notable distinctions emerge within the CVAE framework. Specifically, we notice variations in Pitch Angle and average core speed.

When delving into CVAE with RFT, a more pronounced divergence is evident across multiple variables. Notably, significant differences manifest in Corrected Angle of Attack, Pitch Angle, Core Speed average, and wind speed, underscoring the enhanced performance and nuanced anomaly detection capabilities of the augmented model. This detailed comparison sheds light on how incorporating RFT enriches the model's ability to discern anomalies across a spectrum of variables, thereby enhancing its overall effectiveness in anomaly detection tasks.

\begin{figure}
    \centering
    \includegraphics[width=\columnwidth]{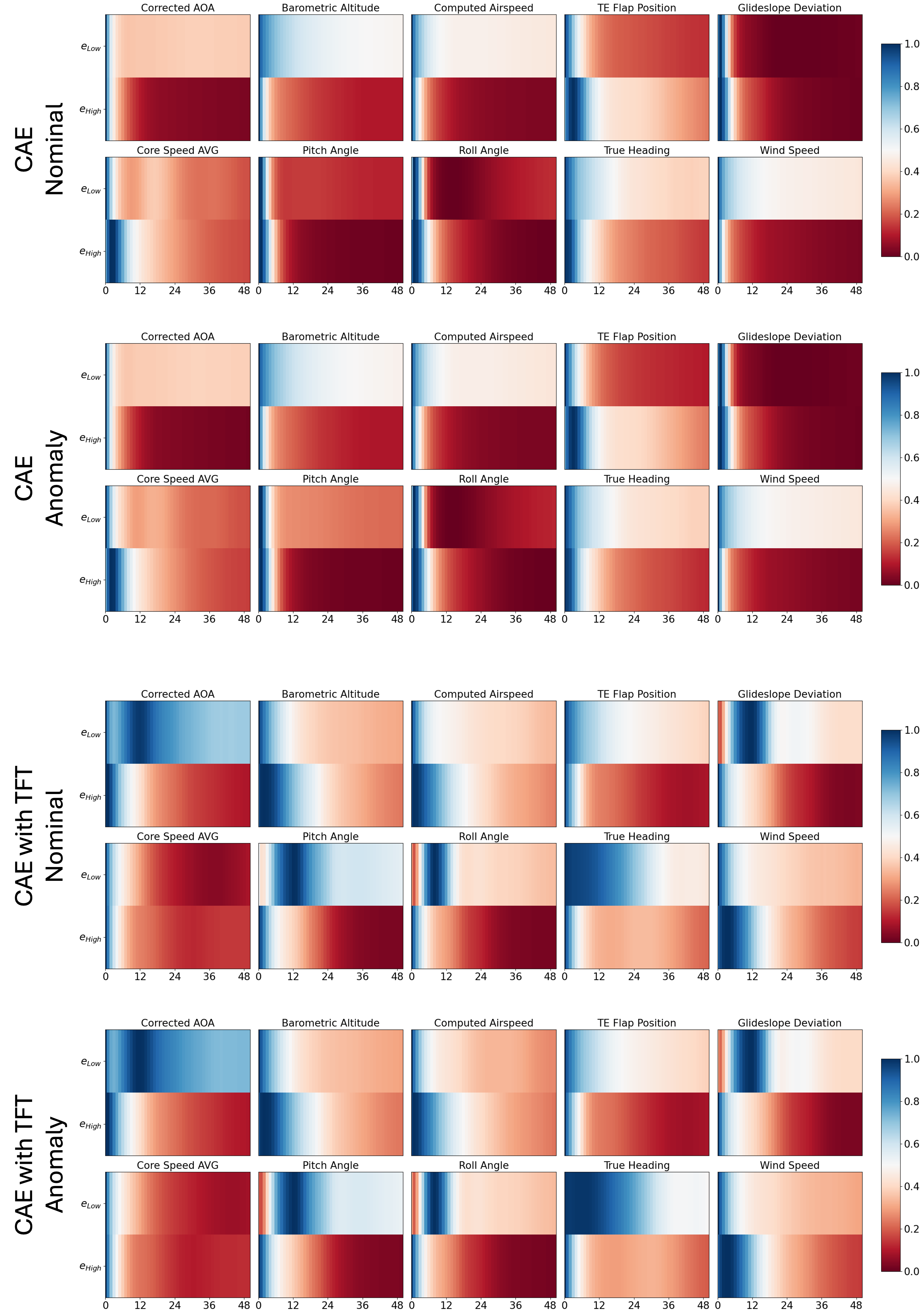}
    \caption{The frequency behavior of CAE and CAE with RFT on nominal and anomaly segments for each variable for Speed anomaly. Each subplot illustrates the separation of low and high frequencies on the y-axis, while the x-axis indicates the training evolution over time.}
    \label{fig:speed_var}
\end{figure}

Following the methodology used in Figures \ref{fig:flaps_var} and \ref{fig:path_var}, we delve into the learning patterns of CAE and CAE with TFT, aiming to discern the nuances between nominal and anomaly cases and gain insights into the pivotal role of variables in anomaly detection. Focusing on the Speed anomaly dataset (Figure \ref{fig:speed_var}), we observe distinct behaviors within the two model configurations.

In the case of CAE, a notable absence of visible differences between nominal and anomaly instances is evident. However, upon closer inspection of CAE with TFT, discernible discrepancies surface, particularly in Computed Airspeed and Pitch Angle. This variance underscores the model's sensitivity to specific variables when equipped with TFT, suggesting that certain anomalies may manifest more prominently in the presence of temporal fusion transformers. Such detailed analysis not only highlights the intricacies of anomaly detection within different model configurations but also underscores the importance of variable selection and model augmentation in enhancing anomaly detection capabilities.

\section{Conclusion}

In this study, we delved into the training and inference improvements of deep neural networks, particularly focusing on Autoencoders (AEs) and Variational Autoencoders (VAEs), using Random Fourier Transformation (RFT). We also explored a trainable variant of RFT (TFT) to understand its impact on model training dynamics and inference capabilities, especially in the context of reconstruction-based anomaly detection.

Our benchmark experiments showcased the evolution of model performance across training epochs, highlighting the effectiveness of incorporating Fourier transformations (both Random and Trainable) in improving convergence and learning sharp features in signals. We employed Frequency-Principle analysis to gain insights into how neural networks interact with different frequency components during training, revealing that networks with Fourier transformations exhibit simultaneous learning of low and high frequencies, mitigating the spectral bias phenomenon observed in conventional DNNs.

Moving to reconstruction-based anomaly detection using high-dimensional datasets, our results demonstrated the superior performance of models with Fourier transformations (RFT and TFT) over vanilla neural networks, particularly in terms of precision, recall, and F1-score metrics. However, we found that the performance gains between Random and Trainable Fourier components were minimal, suggesting limited benefits of training the parameters in the Fourier Transformation process.

In conclusion, our experiments highlight the significant impact of Fourier transformations in enhancing model learning dynamics and improving anomaly detection performance. While both Random and Trainable Fourier transformations show promise, the trainable variant does not exhibit substantial advantages over the random counterpart in our experiments. Future research could delve deeper into the mechanisms behind trainable Fourier transformations and explore additional strategies to further optimize model training and inference processes in deep neural networks.

\section*{Acknowledgment}
This work was funded by NASA’s System-Wide Safety Project under NASA’s Aeronautics Mission Directorate’s Airspace Operations and Safety Program.

\bibliographystyle{plainnat}
\bibliography{biblo}

\section{Appendix}\label{app}

\end{document}